\documentclass[10pt,twocolumn,letterpaper]{article}

\usepackage[pagenumbers]{cvpr} 

\usepackage[table,dvipsnames]{xcolor}
\definecolor{yellow}{rgb}{1, 1, 0.7}
\definecolor{orange}{rgb}{1, 0.85, 0.7}
\definecolor{tablered}{rgb}{1, 0.7, 0.7}

\definecolor{cvprblue}{rgb}{0.21,0.49,0.74}
\usepackage[pagebackref,breaklinks,colorlinks,citecolor=cvprblue]{hyperref}
\usepackage{adjustbox}
\usepackage{amsmath}
\usepackage{algorithm}
\usepackage{algpseudocode}
\usepackage[table]{xcolor}

\title{Improving Multi-View Reconstruction via Texture-Guided Gaussian-Mesh Joint Optimization}

\author{Zhejia Cai\textsuperscript{1,†} \and Puhua Jiang\textsuperscript{1,2,†} \and Shiwei Mao\textsuperscript{1} \and Hongkun Cao\textsuperscript{2,‡} \and Ruqi Huang\textsuperscript{1,‡}
    \\
    \small \textsuperscript{1} SIGS, Tsinghua University
    \quad
    \textsuperscript{2} Peng Cheng Laboratory
    \quad
    † Equal Contribution
    \quad
    ‡ Corresponding Author
    \quad}

\begin{document}
\maketitle
\begin{abstract}
Reconstructing real-world objects from multi-view images is essential for applications in 3D editing, AR/VR, and digital content creation. Existing methods typically prioritize either geometric accuracy (Multi-View Stereo) or photorealistic rendering (Novel View Synthesis), often decoupling geometry and appearance optimization, which hinders downstream editing tasks. 
This paper advocates an unified treatment on geometry and appearance optimization for seamless Gaussian-mesh joint optimization. 
More specifically, we propose a novel framework that simultaneously optimizes mesh geometry (vertex positions and faces) and vertex colors via Gaussian-guided mesh differentiable rendering, leveraging photometric consistency from input images and geometric regularization from normal and depth maps. 
The obtained high-quality 3D reconstruction can be further exploit in down-stream editing tasks, such as relighting and shape deformation. 
Our code will be released in \url{https://github.com/zhejia01/TexGuided-GS2Mesh}
\end{abstract}    
\section{Introduction}

Reconstruction of real-world objects from multi-view images plays a central role in a wide realm of applications, including 3D editing\cite{kuang2023stanford}, AR/VR\cite{bian2023porf,isprs-archives-XLIII-B2-2021-313-2021}, film industry\cite{eisert2020hybrid}, to name a few. 
Upon the recent advances on high-quality reconstruction, in this paper we investigate a relatively underexplored problem -- \emph{how to ease editing operations on both geometry and appearance of digitizations in a unified manner}?
In fact, this problem is becoming increasingly critical with the rapid advancement of interactive virtual environments.
For instance, one might expect to deform an object and/or change lighting condition during interaction.

The key bottleneck of the aforementioned task, in our opinion, is the separated focus of the mainstream 3D representations utilized in reconstruction. 
For instance, classical multi-view stereo (MVS) approaches~\cite{Snavely2006, furukawa2009accurate,Lowe2004,Schonberger2016,Kutulakos2000,Newcombe2011} primarily focus on reconstructing dense point clouds from triangulation guided by photometric consistency and leave appearance alignment to post-processing(\emph{e.g., } texture baking~\cite{furukawa2009accurate}). 
Such approaches can capture fine geometric details while suffering from oversimplified/inconsistent texture maps due to their heavy reliance on geometric priors\cite{yao2020blendedmvs,tatarchenko2019single}. 
On the other hand, Neural View Synthesis (NVS) methods\cite{mildenhall2021nerf,muller2022instant,barron2021mip,yu2024mip,Kerbl20233DGS} have gained considerable popularity in computer vision, which predominantly focus on producing high-fidelity novel view renderings. 
Mesh reconstruction approaches(\emph{e.g., }\cite{Huang20242DGS,Yu2024GaussianOF,wolf2024gsmesh,turkulainen2024dnsplatter} )based on these NVS methods essentially rely on signed distance field(SDF)\cite{osher2004level} representation for geometry extraction and appearance association. 
However, SDF is not trivial to plug into existing geometry processing tools, rendering its difficulty in the geometric editing. 


Perhaps the most relevant works to ours is NVdiffrec(mc)~\cite{Munkberg_2022_CVPR,hasselgren2022shape} and NerF2Mesh~\cite{tang2022nerf2mesh}, which both extract meshes from NVS reconstruction for consequent refinement. 
To attach appearance, these works train neural networks such as coordinate-field MLP~\cite{tang2022nerf2mesh} to address the challenging  problem of mapping texture onto meshes. 
From this point of view, the geometry and appearance remain disentangled in optimization (or learning), hindering their utilities in the scenarios requiring simultaneously editing from both perspectives, as mentioned in the beginning.


To address this problem, our key insight is to \emph{enhance the coherence between geometry and appearance}, in both representation and optimization. 
More specifically, starting from a set of multiview images, we first leverage the recent advances in 3DGS~\cite{Kerbl20233DGS} to achieve appearance reconstruction and extract a coarse mesh. 
Crucially, we advocate to decorate this mesh with \emph{per-vertex color}, which is also accessible from the 3DGS reconstruction. 
Thus, we can optimize geometry and appearance in a unified manner and easily adopt methods developed in geometry processing.  
In particular, we adopt the iterative, inverse-rendering-based remeshing method~\cite{palfinger2022continuous} into our framework. 
Unlike ContinousRemeshing~\cite{palfinger2022continuous} depending on the \emph{ground-truth} normal and depth rendered from the given target geometry, our method can effectively refine the initial mesh via photometric consistency, weak geometric supervision from the initial mesh and some mild geometric regularization. 

Though our approach seems conceptually simple, we need to overcome the disadvantage of per-vertex color encoding. 
More specifically, due to the linear nature of our color coding, it is prone to produce color artifacts, especially around the regions consisting of smooth geometric change but dramatic texture variation. 
To this end, we further propose a Texture-based Edge Length Control (TELC) scheme to robustify our remeshing pipeline. 

Finally, to fully exploit the high-quality textured mesh, we further propose a vertex-Gaussian binding scheme, so that the improved geometry can be transferred to the bound Gaussian, which enables simultaneous material and geometric editing of the reconstructed object.


We conduct a rich set of experiments to verify the effectiveness and efficiency of our pipeline, highlighting its superiority in geometric accuracy, rendering fidelity, relighting precision, and deformation consistency.


\section{Related Work}
\label{sec:relatedwork}

\subsection{Surface Reconstruction with Volume Rendering}
Neural Radiance Fields (NeRF) \cite{mildenhall2021nerf} represent a scene as a continuous volumetric function using a neural network that predicts the color and density for points in 3D space, enabling photo-realistic novel view synthesis. 
3D Gaussian Splatting (3DGS)\cite{Kerbl20233DGS} optimizes an explicit representation through differentiable rasterization, which not only significantly enhances training speed but also improves the quality of novel view synthesis. 

However, NeRF and 3DGS are not specifically designed for mesh extraction tasks, and therefore extracting meshes based on the density of sampled points leads to inaccurate reconstruction results. 
To address these limitations, NeuS\cite{Wang2021NeuSLN} represents surfaces as the zero-level set of SDF and introduces a new volume rendering formulation to reduce geometric bias inherent in conventional volume rendering. 
NeuS2\cite{Wang2022NeuS2FL} and Neuralangelo\cite{Li2023NeuralangeloHN} integrate multi-resolution hash encodings and accelerate training. Methods like IRON~\cite{zhang2022iron}, NeMF~\cite{zhang2023nemf}, Neural Microfacet~\cite{mai2023neural}, and ROSA~\cite{kaltheuner2025rosa} further study object-centric inverse rendering to jointly recover geometry and materials/appearance from images, complementing scene-level surface reconstruction methods. In terms of explicit mesh extraction meshod, SuGaR\cite{Gudon2023SuGaRSG} and Gaussian Surfels\cite{Dai2024HighqualitySR} regulate Gaussians and extract meshes by Poisson reconstruction\cite{kazhdan2006poisson} technique. 
2D Gaussian Splatting (2DGS)\cite{Huang20242DGS} improves upon 3DGS by using 2D oriented planar Gaussian disks and employs TSDF fusion\cite{curless1996volumetric}. Furthermore, Gaussian Opacity Field (GOF)\cite{Yu2024GaussianOF} provides a tetrahedron grid-based technique based on DMTet\cite{shen2021deep} instead of Poisson reconstruction and TSDF fusion. Planar-based Gaussian Splatting Reconstruction (PGSR)~\cite{chen2024pgsr} presents a representation for efficient and high-fidelity surface reconstruction from multi-view RGB images and surpasses all existing methods.
However, solely relying on parameter extraction of meshes from 3D representations can lead to a gap between 2D and 3D representations. 
That is, detailed information in multi-view images may be lost in the process from 2D images to 3D representations to 3D meshes. Therefore, we propose a method that optimizes meshes by simultaneously utilizing 2D images and 3D representations, enabling the meshes to have finer details.
\begin{figure*}[!t]
    \begin{center}
    \includegraphics[width=1\textwidth]{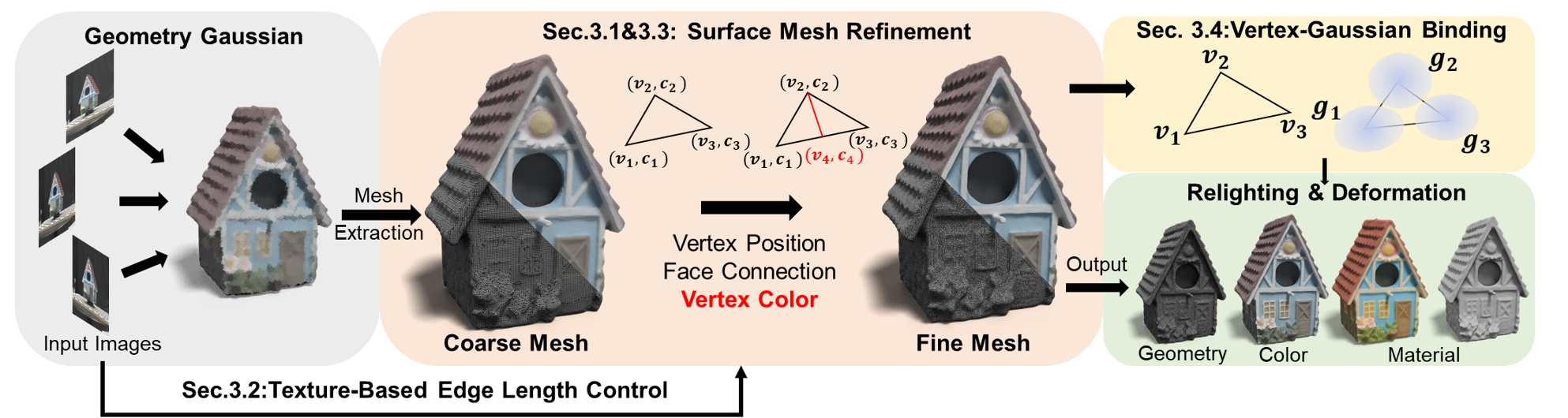} 
    \end{center}

    \caption{The schematic illustration of our pipeline. }\label{fig:pipeline}
\end{figure*}
\subsection{Hybrid of Gaussian Splatting and Mesh Representations}

Recent works in the field of computer graphics and geometry processing have explored hybrid methods that combine the advantages of mesh representations with the flexibility of Gaussian splatting. These approaches typically bind Gaussians to the vertices or faces of a coarse mesh, allowing the Gaussians to benefit from the geometric structure provided by the mesh. The primary goal of these methods is to enhance the rendering quality of the Gaussians, leveraging the mesh's shape to improve the appearance and coherence of the splatting process.

For instance, \textit{Mani-GS} \cite{Gao2024ManiGSGS} presents a hybrid approach that binds Gaussians to a coarse mesh, aiming to refine their appearance through optimization techniques. The idea is to optimize the Gaussian parameters (such as position, scale, and opacity) while keeping the Gaussians aligned with the mesh structure. Similarly, \textit{Gaussian Mesh Splatting} \cite{Gao2024MeshbasedGS} also explores the fusion of Gaussian splatting with mesh representations, primarily focusing on how to deform Gaussians in accordance with mesh transformations, thereby enabling dynamic scene rendering and deformation.

However, these existing methods predominantly focus on binding Gaussians to a static or deformed mesh structure and optimizing their rendering effects. While they effectively improve the visual quality of Gaussians on the mesh, they largely neglect the reverse conversion—how to transfer learned Gaussian attributes back to the mesh for tasks such as relighting or deformation.


\subsection{Reconstruction via Optimization}


Recent works have sought to bridge the gap between implicit neural representations and explicit 3D meshes for practical applications. Among these, NeRF2Mesh\cite{tang2022nerf2mesh} extracts a coarse mesh and iteratively refines both vertex positions and face density using re-projection errors to guide adaptive surface optimization. However, NeRF2Mesh decouples geometry and view-dependent appearance, processing them independently, which may limit the potential for joint optimization. Continuous remeshing\cite{palfinger2022continuous} provides a tool for achieving higher-quality geometric optimization, avoiding unreasonable face patches during the refinement process. Similar to NeRF2Mesh, it also neglects the influence of appearance on the refinement process, potentially missing opportunities for enhanced reconstruction quality. Other end-to-end pipelines \cite{shen2023flexible,son2024dmesh,son2024dmesh++,liu2024gshell,shen2021deep} for mesh reconstruction from multi-view images still face challenges in reconstructing fine details. Like the aforementioned methods, these approaches often disregard the role of appearance, limiting their ability to leverage joint geometry-appearance optimization for improved results.

\section{Method}

As shown in Fig.~\ref{fig:pipeline}, starting from multi-view images, we first use off-the-shelf  3DGS methods~\cite{Huang20242DGS,Yu2024GaussianOF,Kerbl20233DGS} to reconstruct the scene, then compute TSDF upon the 3DGS representation, and finally obtain the initial mesh $\mathcal{M}_{ini} = (V^0, T^0, C^0)$ with marching cube algorithm. Here $V^0 = \{v_i\in \mathbb{R}^3\}_{i= 1}^n$, $T^0 = \{t_j\}_{j = 1}^m$, and $C^0 = \{c_i\in \mathbb{R}^3\}_{i= 1}^n$, are respectively the vertex, face and per-vertex color extracted from the reconstruction. 
From Sec.~\ref{sec:3.1} to Sec.~\ref{sec:3.3}, we introduce our texture-guided remeshing, which effectively refines the geometry of $\mathcal{M}_{ini}$ while preserving rendering quality. 
On top of the improved textured mesh, in Sec.~\ref{sec:3.4} we propose a novel approach to bind mesh to Gaussian, which improves results in tasks such as relighting and deformation.



\subsection{Geometry-Color Remeshing Operations}\label{sec:3.1}

It is well-known that geometry is generally not well reconstructed by 3DGS  on their own (see also Fig.~\ref{fig:vis}). 
Our first goal is to refine $\mathcal{M}_{ini}$.  
Obviously, independently optimizing it with respect to geometric loss would fall short of preserving the rendering quality.  
Our key insight is to introduce the appearance attributes, namely, \emph{color}, to join the geometric refinement. 

For mesh refinement, we adopt the framework of ContinuousRemeshing~\cite{palfinger2022continuous}, which leverage inverse rendering technique~\cite{Laine2020diffrast} to remesh a sphere to a target mesh. 
The remeshing is performed by enforcing the normal and depth image of remeshed object to approximate those computed on the target from multiple views. 
In particular, to accommodate the color attributes, we extend the standard remeshing operations to the following geometry-color-based ones: 

\noindent\textbf{Edge Split with Color Interpolation}: When splitting edge $e = (v_i, v_j)$ on triangle $(v_i, v_j, v_l)$, we create a new vertex $v_k$ with position and color bilinearly interpolated at the midpoint of $e$, after that creating three edges $e_1,e_2,e_3$ and removing one edge $e$:
\begin{equation}
\begin{aligned}
&(v_k, c_k) := \left(\frac{v_i+v_j}{2}, \frac{c_i+c_j}{2}\right), \\
&e_1 := (v_l,v_k), e_2 := (v_i,v_k), e_3 := (v_j,v_k), \\
&\text{remove} \, e = (v_i,v_j).
\end{aligned}
\end{equation}

\noindent\textbf{Edge Collapse with Color Fusion}: Collapsing edge $e = (v_i, v_j)$ propagates color information through merging the two endpoints of the edge to the midpoint.We move $v_i$ to the midpoint and still mark it as $v_i$. For any edge $e_{any} = (v_{any},v_j)$ connected to vertex $v_j$, we change it to $(v_{any},v_i)$. We define all edges between two endpoints with more than one edge as redundant and remove them:
\begin{equation}
\begin{aligned}
&(v_i, c_i) := \left(\frac{v_i+v_j}{2}, \frac{c_i+c_j}{2}\right), \\
&e_{any} = (v_{any},v_j) \rightarrow (v_{any},v_i), \\
&\text{remove} \, e \,\text{where redundant}.
\end{aligned}
\end{equation}

\noindent\textbf{Edge Flip with Color Preservation (optional)}: For edge $e = (v_i, v_j)$ between triangles $(v_i, v_j, v_k)$ and $(v_i, v_j, v_l)$, flipping to $(v_k, v_l)$ preserves color coherence through:
\begin{equation}
\begin{aligned}
&e = (v_i, v_j) \rightarrow (v_k, v_l).
\end{aligned}
\end{equation}
To preserve color consistency during optimization, we note that edge flipping can introduce abrupt color changes at patch centroids due to interpolation, particularly when neighboring faces exhibit significant color variations. Therefore, we implement edge flipping intermittently, executing the operation every few optimization steps rather than continuously.

We defer details of our optimization goal, which involves photometric consistency and geometric regularization, to Sec.~\ref{sec:3.3}. Similarly, we refer readers to Sec.~\ref{sup:remesh} of the Supp. for the details of the remesh algorithm.

\subsection{Texture-Based Edge Length Control}\label{sec:3.2}

\begin{figure}[t!]
  \begin{center}
\includegraphics[width=0.5\textwidth]{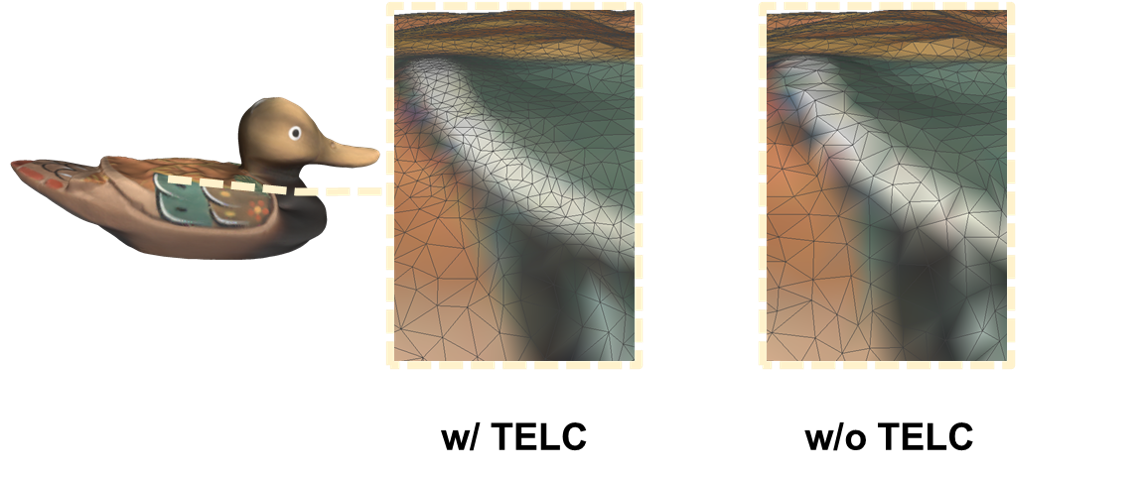}
  \end{center}
  \vspace{-8mm}
\caption{Remeshing results with (middle) and without (right) texture density based edge length control (\emph{i.e., }TELC). }
\label{fig:fft}

\end{figure}

Though the geometry-color remeshing operations presented in the last part enables flexible and efficient update on color attributes of each vertex, it can potentially introduce color artifacts due to the linear nature of color assignment. 
Therefore, we shall take gradients over the appearance domain into consideration of performing remeshing operations. 

To see this, let us consider the mallard on the left of Fig.~\ref{fig:fft}, whose wing exhibits both sharp color transition (from green to white) and smooth geometric change.  
Without control signal from appearance domain, we end up at the right panel of Fig.~\ref{fig:fft} -- there exists large triangle face crossing the boundary, leading to color leakage in such areas. 
Intuitively, we would like to have smaller triangles crossing the boundary in appearance, and respectively larger triangles among flat regions from the perspectives of both geometry and appearance. 
To achieve such, we introduce a simple yet effective edge length control scheme, which incorporates the frequency change computed in the appearance domain. 

In ContinousRemeshing~\cite{palfinger2022continuous}, one computes a constant optimal edge length $l_{ref}^{k+1}$ at each iteration based on the geometry obtained at $k^{th}$ iteration. 
Subsequently, we define edge length tolerance \(\epsilon\) to constrain the range of edge length. For each edge $l$ at ${(k+1)}^{th}$ iteration, one performs remeshing if its length deviates from $l_{ref}^k$ by a margin, namely, when 
\begin{equation}\label{eqn:cond}
	|length(l) - l_{ref}^{k+1}| > \epsilon \times l_{ref}^{k+1}.
\end{equation}



Now we let $\mathcal{M}$ be the mesh at the ${(k+1)}^{th}$ iteration of remeshing, and computes $l_{ref}^{k+1}$ following~\cite{palfinger2022continuous}, which is based on geometry. 
Recall that we are given multi-view images $\mathcal{I} = \{I_1, I_2, ..., I_s\}$ as input, where each $I_i\in \mathbb{R}^{3\times H\times W}$. 
We proceed through the following steps:

\noindent\textbf{Compute texture density map:} For each pixel, $p$ in $I_i$, we consider the $3\times 3$ neighborhood around it, then perform Fast Fourier transform (FFT) and compute the magnitude of the FFT output, which is a single scalar value assigned to $p$, reflecting the oscillation in the regarding patch.
Going through all pixels and all images, we obtain the texture density map $\mathcal{F} = \{f_1, f_2, ..., f_s\}$, where each $f_i: p\in I_i\rightarrow \mathbb{R}^+$. 

\noindent\textbf{Back-project texture density map to meshes and normalize: }
For each vertex $v_p$ in the mesh $\mathcal{M}$, we consider the image subset where it is visible and denote by $vis(p)$ the regarding indices in $\{1, 2, ..., s\}$. 
We back-project $v_p$ to a pixel in each $I_j, j\in vis(p)$, and let $f_j(p)$ be the texture density of the very pixel in $f_j$. 
The texture density of $p$ is then defined as
\begin{equation}
	f(v_p) = \frac{\sum_{j\in vis(p)} f_j(p)}{\# vis(j)}, 
\end{equation}
where $\#A$ returns the cardinality of set $A$. 
Finally, we normalize the per-vertex texture density as follows, so that $0\leq f(v_p)\leq 1$ for all $p$ 
\begin{equation}
	f(v_p) \leftarrow \frac{f(v_p)-\min\{f(v_q), v_q\in \mathcal{M}\}}{\max\{f(v_q), v_q\in \mathcal{M}\} - \min\{f(v_q), v_q\in \mathcal{M}\}}.
\end{equation}

\noindent\textbf{Compute per-edge texture density map: }
Now considering an edge of $\mathcal{M}$, $l = (v_p, v_q)$, we set the texture density of $l$ as
\begin{equation}
	F_l = (f(v_p) + f(v_q))/2.
\end{equation}

Our adaptive edge control scheme then injects the per-edge texture density into Eqn.~\ref{eqn:cond}, namely, we perform remeshing on edge $l$ whenever
\begin{equation}
	|length(l) - l_{ref}^{k+1}\times (1 - F_l)| > \epsilon \times l_{ref}^{k+1}\times (1-F_l).
\end{equation} 
Intuitively, when $l$ is among a region with high frequency, $1-F_l$ approaches $0$, which makes it easier to trigger the remeshing condition. 

\definecolor{tablered}{RGB}{255,200,200} 

\begin{table*}[!t]
\caption{\centering Quantitative comparison on the DTU Dataset. Our method largely improves the reconstruction accuracy on other explicit mesh reconstruction methods with a short refinement process.}
\label{table:dtugeo}
\centering
\resizebox{\textwidth}{!}{
\begin{tabular}{l c c c c c c c c c c c c c c c | c | c}
\toprule
\textbf{Method} & 24 & 37 & 40 & 55 & 63 & 65 & 69 & 83 & 97 & 105 & 106 & 110 & 114 & 118 & 122 & Mean & Time(hours) \\
\midrule
\textbf{NeuS}
& 1.00 & 1.37 & 0.93 & 0.43 & 1.10 & \cellcolor{yellow}{0.65} & 0.57 & 1.48 & 1.09 & 0.83 & \cellcolor{yellow}{0.52} & 1.20 & 0.35 & 0.49 & 0.54 & 0.84 & $>12$ \\
\textbf{Neuralangelo}
& \cellcolor{orange}{0.37}
& \cellcolor{yellow}{0.72}
& \cellcolor{yellow}{0.35}
& \cellcolor{yellow}{0.35}
& \cellcolor{yellow}{0.87}
& \cellcolor{yellow}{0.54}
& \cellcolor{yellow}{0.53}
& 1.29
& \cellcolor{yellow}{0.97}
& 0.73
& \cellcolor{tablered}{0.47}
& \cellcolor{yellow}{0.74}
& \cellcolor{yellow}{0.32}
& \cellcolor{yellow}{0.41}
& 0.43
& \cellcolor{yellow}{0.61}
& $>12$ \\
\midrule
\textbf{3DGS}
& 1.45 & 1.46 & 1.85 & 1.47 & 2.56 & 2.19 & 1.26 & 1.93 & 1.73 & 1.51 & 1.69 & 2.04 & 1.19 & 1.09 & 1.10 & 1.63 & 0.2 \\
\textbf{Ours + 3DGS}
& \underline{1.25} & \underline{1.32} & \underline{1.53} & \underline{1.03} & \underline{2.55} & \underline{2.05} & \underline{1.09} & \underline{1.81} & \underline{1.59} & \underline{1.42} & \underline{1.46} & 2.04 & \underline{0.96} & \underline{0.81} & \underline{0.86} & \underline{1.45} & 0.2 (+0.1) \\
\midrule
\textbf{GOF}
& 0.47 & 0.82 & 0.40 & 0.36 & 1.28 & 0.83 & 0.76 & 1.19 & 1.24 & 0.75 & 0.74 & 1.12 & 0.49 & 0.69 & 0.57 & 0.78 & 0.3 \\
\textbf{Ours + GOF}
& \underline{0.42}
& \underline{0.76}
& \cellcolor{yellow}{\underline{0.35}}
& \cellcolor{tablered}{\underline{0.33}}
& \underline{1.22}
& \underline{0.74}
& \underline{0.66}
& \cellcolor{yellow}{\underline{1.13}}
& \underline{1.23}
& \underline{0.70}
& \underline{0.65}
& 1.14
& \underline{0.43}
& \underline{0.57}
& \underline{0.46}
& \underline{0.72}
& 0.3 (+0.1) \\
\midrule
\textbf{2DGS}
& 0.49 & 0.82 & \cellcolor{orange}{0.34} & 0.42 & 0.95 & 0.86 & 0.82 & 1.29 & 1.24 & 0.66 & 0.64 & 1.44 & 0.41 & 0.67 & 0.50 & 0.77 & 0.2 \\
\textbf{Ours + 2DGS}
& \cellcolor{yellow}{\underline{0.40}}
& \underline{0.75}
& \cellcolor{tablered}{\underline{0.30}}
& \cellcolor{tablered}{\underline{0.33}}
& 0.96
& \underline{0.76}
& \underline{0.71}
& \underline{1.24}
& \underline{1.20}
& \cellcolor{yellow}{\underline{0.60}}
& \underline{0.55}
& \underline{1.40}
& \underline{0.39}
& \underline{0.55}
& \cellcolor{yellow}{\underline{0.40}}
& \underline{0.70}
& 0.2 (+0.1) \\
\midrule
\textbf{PGSR}
& \cellcolor{tablered}{0.34}
& \cellcolor{orange}{0.55}
& 0.40
& 0.36
& \cellcolor{orange}{0.78}
& \cellcolor{orange}{0.57}
& \cellcolor{orange}{0.49}
& \cellcolor{orange}{1.08}
& \cellcolor{orange}{0.87}
& \cellcolor{orange}{0.59}
& \cellcolor{orange}{0.49}
& \cellcolor{orange}{0.51}
& \cellcolor{orange}{0.30}
& \cellcolor{orange}{0.37}
& \cellcolor{orange}{0.34}
& \cellcolor{orange}{0.53}
& 0.6 \\
\textbf{Ours + PGSR}
& \cellcolor{tablered}{0.34}
& \cellcolor{tablered}{\underline{0.50}}
& \underline{0.38}
& \cellcolor{orange}{\underline{0.34}}
& \cellcolor{tablered}{\underline{0.74}}
& \cellcolor{tablered}{\underline{0.54}}
& \cellcolor{tablered}{\underline{0.47}}
& \cellcolor{tablered}{\underline{1.03}}
& \cellcolor{tablered}{\underline{0.85}}
& \cellcolor{tablered}{\underline{0.56}}
& \cellcolor{tablered}{\underline{0.47}}
& \cellcolor{tablered}{\underline{0.49}}
& \cellcolor{tablered}{\underline{0.29}}
& \cellcolor{tablered}{\underline{0.36}}
& \cellcolor{tablered}{\underline{0.33}}
& \cellcolor{tablered}{\underline{0.51}}
& 0.6 (+0.15) \\
\bottomrule
\end{tabular}
}
\end{table*}

\begin{table*}
\caption{\centering Quantitative comparison on the DTC Dataset (values multiplied by 1000). Our method improves the object reconstruction accuracy on baseline methods.}
\label{table:dtcgeo}
\centering
\resizebox{\textwidth}{!}{
    \begin{tabular}{lcccccccccccccccc|c}
    \toprule
    \textbf{Method} & \textbf{Airplane} & \textbf{BirdHouse} & \textbf{Car} & \textbf{CaramicBowl} 
    & \textbf{Cup} & \textbf{DutchOven} & \textbf{Hammer} & \textbf{Keyboard} 
    & \textbf{Kitchen} & \textbf{Mallard} & \textbf{Planter} & \textbf{Pottery} 
    & \textbf{Shoe} & \textbf{Spoon} & \textbf{Teapot} & \textbf{Vase} & \textbf{Mean} \\
    \midrule
    \textbf{GOF} 
        & 3.33   
        & 1.60   
        & \colorbox{orange}{0.89}   
        & 2.83   
        & 2.60   
        & \colorbox{yellow}{1.93}   
        & 1.39   
        & 4.23   
        & 2.77   
        & \colorbox{yellow}{1.60}   
        & 2.52   
        & \colorbox{yellow}{3.44}   
        & 2.13   
        & 2.60   
        & \colorbox{tablered}{{1.62}}   
        & 4.12   
        & 2.48\\ 
    
    \textbf{Ours + GOF} 
        & \underline{2.67}   
        & \underline{1.17}   
        & \colorbox{yellow}{0.91}·   
        & \colorbox{yellow}{\underline{2.77}}   
        & \colorbox{yellow}{\underline{2.30}}   
        & 2.01   
        & \colorbox{yellow}{\underline{0.99}}   
        & \colorbox{tablered}{\underline{{2.23}}}   
        & \underline{2.19}   
        & \colorbox{tablered}{\underline{1.26}}   
        & \underline{2.19}   
        & 3.84   
        & \underline{2.12}   
        & \underline{2.18}   
        & \colorbox{tablered}{\underline{{1.62}}}   
        & \underline{3.51}   
        & \underline{2.12}\\ 
    \midrule
    \textbf{2DGS}
        & 2.24 & 1.26 & 1.06 & 3.28 & 2.84 & \colorbox{yellow}{1.90} & 1.20 & \colorbox{orange}{2.49}
        & 1.67 & 1.99 & 1.64 & 3.79 & 2.20 & 1.86 & 2.29 & \colorbox{yellow}{2.93} & 2.17\\
    
    \textbf{Ours + 2DGS} 
        & \colorbox{yellow}{\underline{2.21}}   
        & \colorbox{orange}{\underline{1.10}}   
        & \colorbox{tablered}{\underline{{0.84}}}   
        & \underline{3.21}   
        & \underline{2.48}   
        & \colorbox{orange}{\underline{1.83}}   
        & \colorbox{orange}{\underline{0.92}}   
        & \colorbox{yellow}{2.50}   
        & \colorbox{yellow}{\underline{1.22}}   
        & \underline{1.63}   
        & \colorbox{yellow}{\underline{1.46}}   
        & 3.93   
        & \colorbox{yellow}{\underline{2.09}}   
        & \colorbox{yellow}{\underline{1.84}}   
        & \colorbox{orange}{\underline{2.21}}   
        & \colorbox{orange}{\underline{2.28}}   
        & \colorbox{yellow}{\underline{1.98}}\\ 

    \midrule
    \textbf{PGSR}
        & \colorbox{tablered}{{1.99}} & \colorbox{yellow}{1.13} & 0.96 & \colorbox{orange}{1.90} & \colorbox{orange}{1.46} & \colorbox{tablered}{{1.40}} & 1.01 & \colorbox{yellow}{2.50}
        & \colorbox{orange}{0.89} & 1.65 & \colorbox{orange}{1.20} & \colorbox{orange}{1.93} & \colorbox{orange}{2.07} & \colorbox{orange}{1.75} & 2.33 & \colorbox{yellow}{2.32} & \colorbox{orange}{1.66}\\
    \textbf{Ours + PGSR}
        & \colorbox{orange}{2.01} & \colorbox{tablered}{\underline{{1.05}}} & \colorbox{yellow}{\underline{0.91}} & \colorbox{tablered}{\underline{{1.78}}} & \colorbox{tablered}{\underline{{1.35}}} & \colorbox{tablered}{{1.40}} & \colorbox{tablered}{\underline{{0.88}}} & \colorbox{yellow}{2.50}
        & \colorbox{tablered}{\underline{{0.80}}} & \colorbox{orange}{\underline{{1.41}}} & \colorbox{tablered}{\underline{{1.06}}} & \colorbox{tablered}{\underline{{1.82}}} & \colorbox{tablered}{\underline{{2.06}}} & \colorbox{tablered}{\underline{{1.63}}} & \colorbox{yellow}{\underline{2.22}} & \colorbox{tablered}{\underline{{1.98}}} & \colorbox{tablered}{\underline{{1.55}}}\\ 
    \bottomrule
    \end{tabular}
}
\end{table*}

With the above scheme, our remeshing result is shown in the middle of Fig.~\ref{fig:fft}, which is clearly improved as the band region are entirely white now. 
Overall, our scheme allows for a more fine-grained control of mesh resolution, especially in regions where textures exhibit high-frequency details, leading to a more accurate and visually consistent result.

\subsection{Mesh Optimization via Inverse Rendering}\label{sec:3.3}

Now we proceed to describe our remeshing procedure. 
We first render pseudo-ground-truth depth maps $\mathcal{D} = \{d_1, d_2, ..., d_s\}$ and normal maps $\mathcal{N} = \{n_1, n_2, ..., n_s\}$ via $\mathcal{M}_{ini}$, the initial mesh extracted from Gaussians, from the input camera views for later regularization. 

At each iteration of remeshing, we denote the regarding mesh $\mathcal{M}^k = (V^k, T^k, C^k)$. 
Via rasterization function $\mathbf{R}$, we can compute 
\begin{equation}
I^k_i, d^k_i, n^k_i = \mathbf{R}(V^k, T^k, C^k, MV_{i}, P_{i}), i = 1, 2, ..., s, 
\end{equation}
where $MV_{i}$ is the model-view matrix of the camera, and $P_{i}$ is the projection matrix of the camera, and $I^k_i, d^k_i, n^k_i$ are respectively the RGB, depth and normal image rendered from viewpoint $i$.

In general, at each iteration, we enforce 1) the RGB rendering to approximate the input multiview images; 2) the rendered depth and normal images to be close to the one computed on $\mathcal{M}_{ini}$ for regularization; 3) the remeshed vertex positions and normals are smooth regarding mesh Laplacian. 

It is worth noting that, though our remeshing pipeline is built on~\cite{palfinger2022continuous}, our loss design differs significantly from the former as 1) We introduce photometric consistency into remeshing; 2) The optimization in \cite{palfinger2022continuous} depends on the \emph{ground-truth} normal and depth of the target, while our framework leverages pseudo-label obtained from multiview images; 3) \cite{palfinger2022continuous} is primarily guided by the ground-truth geometry, therefore it applies Laplacian-based smoothness regularization on the gradient, which is too weak for our challenging task.


To conclude, our loss function is as follows:
\begin{equation}
\mathcal{L} = \lambda_{rgb} \mathcal{L}_{rgb} + \lambda_{geo} \mathcal{L}_{geo} + \lambda_{reg} \mathcal{L}_{reg}, 
\end{equation}
where \( \mathcal{L}_{rgb} \) is the loss term for RGB images, \( \mathcal{L}_{geo} \) is the loss term for depth maps and normal maps, and \( \mathcal{L}_{reg} \) is the regularization term, which includes Laplacian smoothing and mesh normal consistency. 
The coefficients \(\lambda_{rgb}\), \(\lambda_{geo}\), and \(\lambda_{reg}\) are the respective weights for each term. 
We refer readers to Sec.~\ref{sup:loss} of the Supp. for the details of each loss term.


\subsection{Vertex-Gaussian Binding for Relighting and Deformation}\label{sec:3.4}

In this part, we introduce a vertex-Gaussian binding scheme, which exploit the improved geometry we obtained above to enhance down-stream editing applications





Given an optimized mesh \( \mathcal{M}^* = (V^*, T^*, C^*) \),
we define the transformation from the mesh to Gaussian parameters. 
For each vertex \( v_i \in V^* \), we associate a corresponding Gaussian with the following parameters:
\begin{enumerate}
    \item \textbf{Position}: Direct correspondence between vertex \( v_i \) and Gaussian position \( \mu_i \).
    \item \textbf{Scale}: Composed of three components capturing local edge projections on the tangent plane.
    \item \textbf{Rotation}: Orthonormal basis derived from vertex normal and tangent vectors.
    \item \textbf{Opacity}: In our method, we assign a constant opacity value of 0.9 to each Gaussian, assuming that every point on the mesh is visible.
    \item \textbf{Spherical Harmonics (SH) coefficients}: In our method, we assign the low-order SH coefficients directly from the vertex color \( c_i \), and set the higher-order coefficients to zero.
\end{enumerate}

We refer readers to Sec.~\ref{sup:gsbind} of the Supp. for more details. 
In Sec.~\ref{sec:rel}, we feed in the above constructed 3DGS as input to R3DG~\cite{gao2024relightable}, and demonstrate that the improved initialization directly boosts the final relighting performance.

\begin{figure*}[!t]
  \begin{center}
\includegraphics[width=0.95\textwidth]{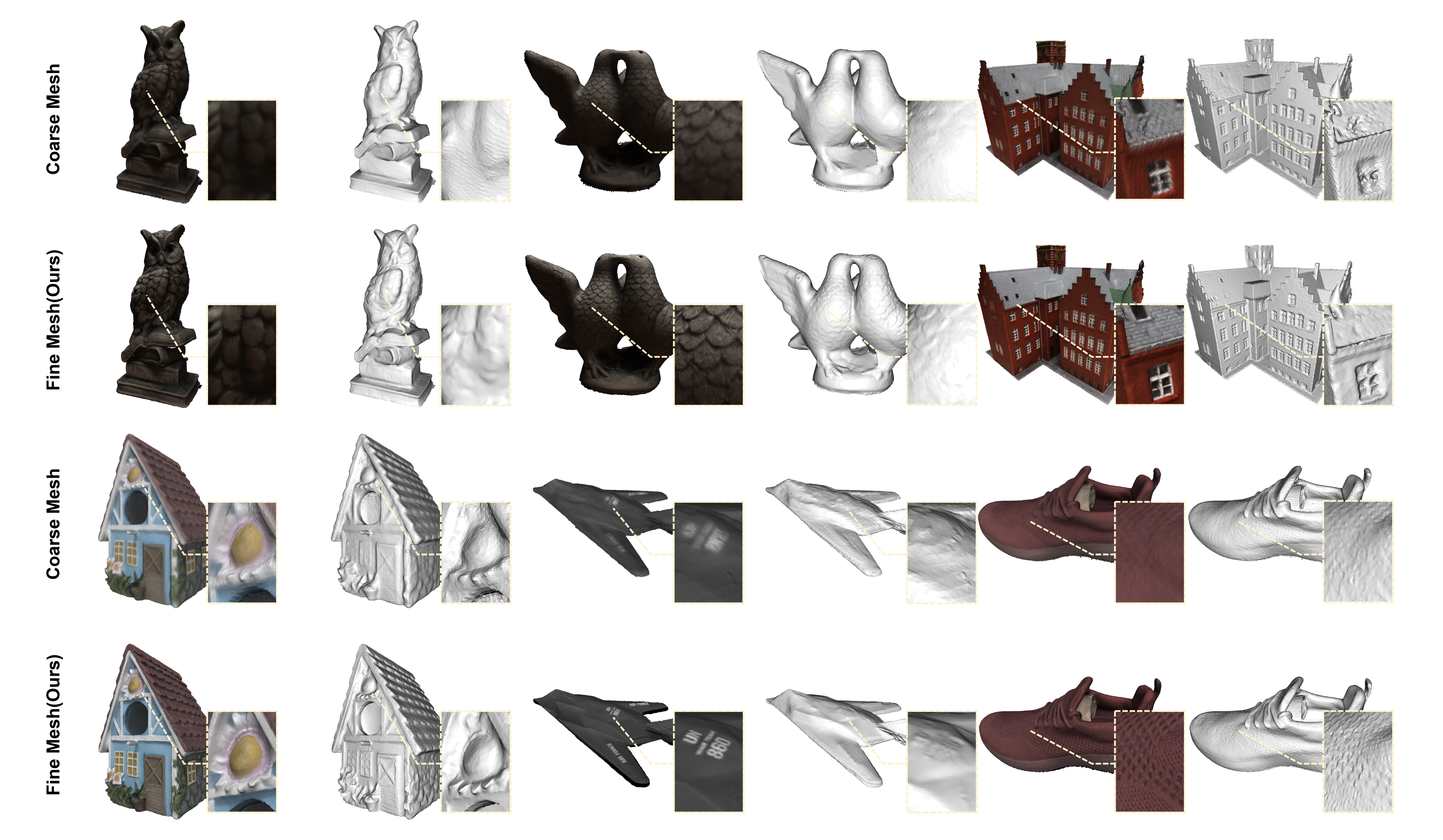}
  \end{center}
  \vspace{-2em}
\caption{Qualitative results on DTU and DTC dataset }
\label{fig:vis}

\end{figure*}

\section{Experiments}
\subsection{Remeshing evaluation}

\noindent\textbf{Baselines:} Our baselines include both implicit ones -- NeuS\cite{Wang2021NeuSLN} and Neuralangelo\cite{Li2023NeuralangeloHN} and explicit ones -- 3DGS\cite{Kerbl20233DGS}, 2DGS\cite{Huang20242DGS}, GOF\cite{Yu2024GaussianOF} and PGSR\cite{chen2024pgsr}. 
Our refinement is mainly applied on the latter four. 

\noindent\textbf{Benchmarks:} 
We evaluate the performance of our method on various datasets. \textbf{DTU}\cite{jensen2014large} dataset comprises 15 scenes, each with 49 or 64 images of resolution 1600$\times$1200. Different from the DTU dataset, which only includes partial surfaces of objects, the latest Digital Twin Catalog(\textbf{DTC})\cite{Pan_2023_ICCV} dataset provides multiview images of complete objects along with ground-truth meshes for evaluation.  DTC dataset contains more than 100 objects and theirs multiview images. We consider 16 cases (\emph{e.g.}, airplane, birdhouse) as our benchmark and down-sample all images in DTC dataset to half of their original size (\emph{i.e.}, 1000$\times$1000).

\noindent\textbf{Implementation Details:} 
For \textbf{DTU}\cite{jensen2014large} dataset, we initialize max edge length to \(1 \mathrm{e}{-3}\) and min edge length to \(1 \mathrm{e}{-4}\). 
For \textbf{DTC} \cite{Pan_2023_ICCV} dataset, we initialize max edge length to \(4 \mathrm{e}{-3}\) and min edge length to \(4 \mathrm{e}{-4}\). 
During surface mesh refinement, we set \(\lambda_{rgb}\) to 3.0, \(\lambda_{reg}\) to 0.3, \(\lambda_{geo}\) to 0.1 and edge length tolerance \(\epsilon\) to 0.5. We train $1000$ iterations per scene and the learning rate is set to \(1 \mathrm{e}{-3}\). We
conduct all the experiments on a single RTX3090 GPU.

\noindent\textbf{Geometry Evaluation:}
We first compare against SOTA implicit and explicit methods on Chamfer Distance and training time using the DTU dataset in Tab.~\ref{table:dtugeo}. 
Our method outperforms all compared methods in terms of Chamfer Distance. 
We integrated our method into 3DGS, GOF, 2DGS and PGSR, respectively, and observe consistent improvement in each case. 
Notably, our method requires only a short optimization time to improve the quality of surface reconstruction, making it plug-and-play for any Gaussian-based surface reconstruction method. 
As illustrated in Fig.~\ref{fig:vis}, the surfaces reconstructed by 2DGS exhibit geometric blurring , while our approach can achieve higher quality reconstruction results. 
We further compare against  2DGS, GOF and PGSR on \textbf{DTC} dataset in Tab.~\ref{table:dtcgeo}. 
The same trend is observed -- As shown in Fig.~\ref{fig:vis}, our method maintains excellent performance in areas with intricate geometric details, particularly evident in the geometry of the athletic shoe at the bottom right corner of the image, where the geometric intricacies of the shoe's surface is better recovered by our refiment.

\noindent\textbf{Rendering Evaluation: }
To evaluate the quality of mesh vertex color, we render the mesh to pixel space and compare rendered image with ground-truth image. As shown in Tab.~\ref{table:rendering}, our method achieves a significant improvement in rendering quality compared to coarse meshes extracted from various Gaussian Splatting (GS) approaches. 
More specific details are illustrated in Fig.~\ref{fig:vis}: the rendering results of coarse meshes exhibit blurriness and a lack of detail clarity. 
After our refinement, texture details such as the text on the airplane, mesh surface details of the sneakers, and window details of the house in the figure have been fully restored.
\begin{table}[tbp]
    \centering
    \caption{Quantitative comparison on DTU and DTC Dataset}
    \label{table:rendering}
    \begin{adjustbox}{max width=0.48\textwidth}

    \begin{tabular}{l|ccc|ccc}
    \toprule
    & \multicolumn{3}{c|}{\textbf{DTU objects}} 
    & \multicolumn{3}{c}{\textbf{DTC objects}} \\
    \cline{2-7}
    \textbf{Method} 
    & PSNR~$\uparrow$ & SSIM~$\uparrow$ & LPIPS~$\downarrow$ 
    & PSNR~$\uparrow$ & SSIM~$\uparrow$ & LPIPS~$\downarrow$ \\
    \midrule
    GOF            
    & 24.81 & 0.858 & 0.194 
    & 25.16 & 0.949 & 0.063 \\
    Ours + GOF
    & \underline{25.63} & \underline{0.897} & \underline{0.160}
    & \underline{\textbf{27.12}} & \underline{0.960} & \underline{0.049} \\
    \midrule
    2DGS           
    & 23.82 & 0.853 & 0.199 
    & 25.16 & 0.950 & 0.058 \\
    Ours + 2DGS    
    & \underline{\textbf{26.21}} & \underline{\textbf{0.906}} & \underline{\textbf{0.148}} 
    & \underline{26.25} & \underline{\textbf{0.962}} & \underline{\textbf{0.042}} \\
    \bottomrule
    \end{tabular}
    \end{adjustbox}
\end{table}

\subsection{Relighting and Deformation Editing}\label{sec:rel}

As mentioned in Sec.~\ref{sec:3.4}, with our optimized meshes with vertex colors, we initialize Gaussian Splatting (GS) using our binding scheme and feed such as input to R3DG~\cite{gao2024relightable} to learn material parameters. 
The learned material parameters are transferred to the mesh via Gaussian binding correspondences, with a 100-iteration noise filtering applied during backpropagation to mitigate renderer discrepancies.
We demonstrate the effectiveness of our method by performing relighting on the Synthetic4Relight dataset\cite{Zhang_2022_CVPR}.

\begin{figure}[t!]
  \begin{center}
\includegraphics[width=0.5\textwidth]{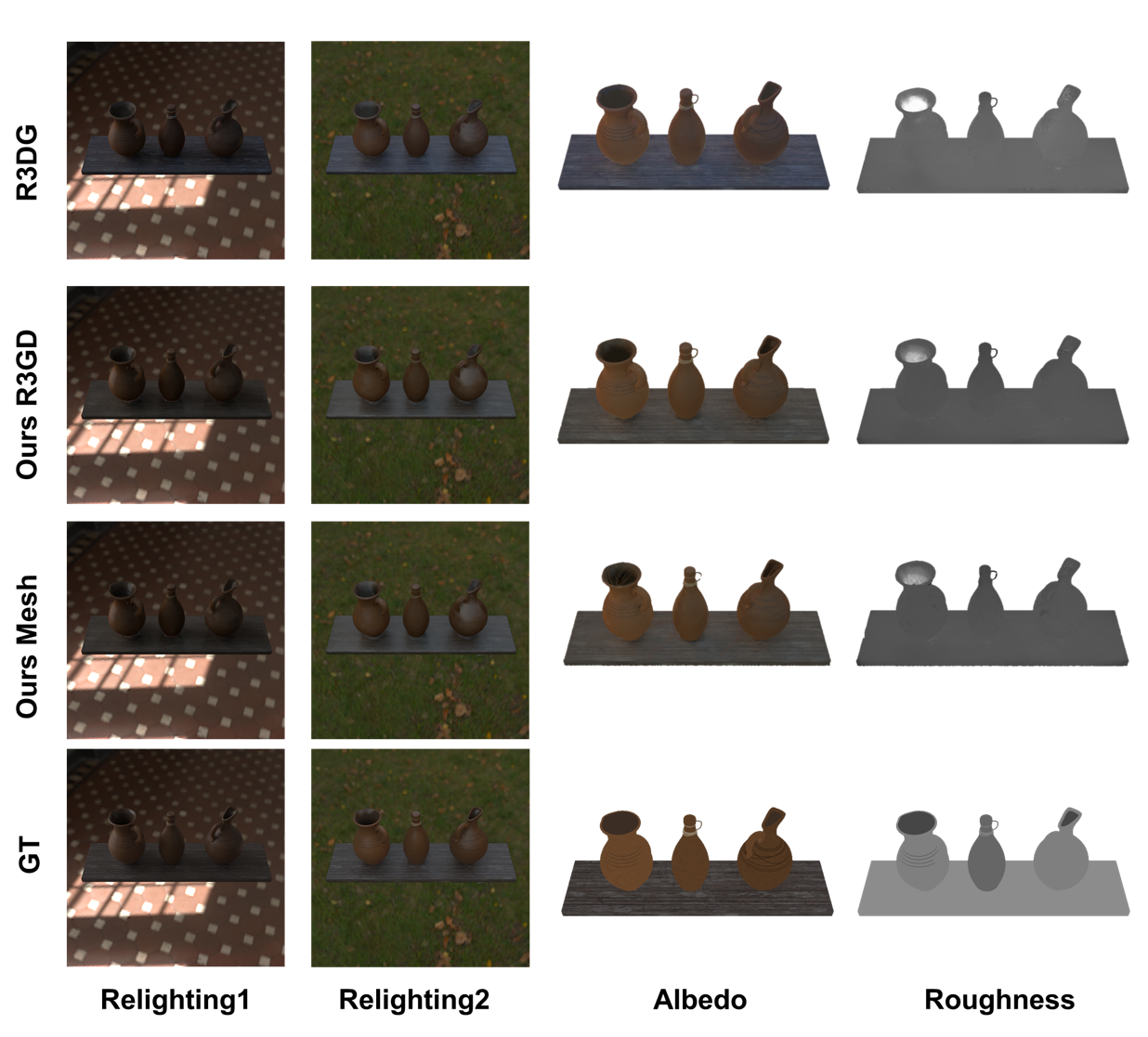}
  \end{center}
\vspace{-2em}
\caption{Qualitative results on Synthetic4Relight dataset }
\label{fig:relight}
\end{figure}
\begin{table*}[htbp]
    \centering

    \caption{Quantitative Results on Synthetic Dataset}
    \label{tab:relight}
    \resizebox{\textwidth}{!}{
    \begin{tabular}{l|ccc|ccc|ccc|c|c}
    \toprule
    & \multicolumn{3}{c|}{\textbf{Novel View Synthesis}} 
    & \multicolumn{3}{c|}{\textbf{Relighting}} 
    & \multicolumn{3}{c|}{\textbf{Albedo}} 
    & \textbf{Roughness} 
    & \textbf{Time} \\
    \cline{2-12}
    \textbf{Method} 
    & PSNR~$\uparrow$ & SSIM~$\uparrow$ & LPIPS~$\downarrow$ 
    & PSNR~$\uparrow$ & SSIM~$\uparrow$ & LPIPS~$\downarrow$ 
    & PSNR~$\uparrow$ & SSIM~$\uparrow$ & LPIPS~$\downarrow$ 
    & MSE~$\downarrow$ & (hours) \\
    \midrule

    R3DG
    &36.80 &0.982 &0.028
    &31.00 &0.964 &0.050
    &28.31 &0.951 &0.058
    & 0.013 & 1.5
    \\
    Ours R3DG
    &33.44 & 0.969 & 0.052
    & 32.87 & 0.965 & 0.054
    & 29.20 & 0.948 & 0.065 
    & 0.009 & 1 \\

    Nvdiffrecmc
    &34.29 &0.967 &0.068
    &24.22 &0.943 &0.078
    &29.61 &0.945 &0.075
    &0.009 &4.17
    \\
    Ours Mesh
    &31.36 &0.962 & 0.055
    & 30.40 & 0.942 & 0.083 
    & 27.35 & 0.928 & 0.081 
    & 0.010 & 1+2m \\

    \bottomrule
    \end{tabular}
    }\label{table:relight}
\end{table*}

\begin{figure}[t!]
  \begin{center}
\includegraphics[width=0.5\textwidth]{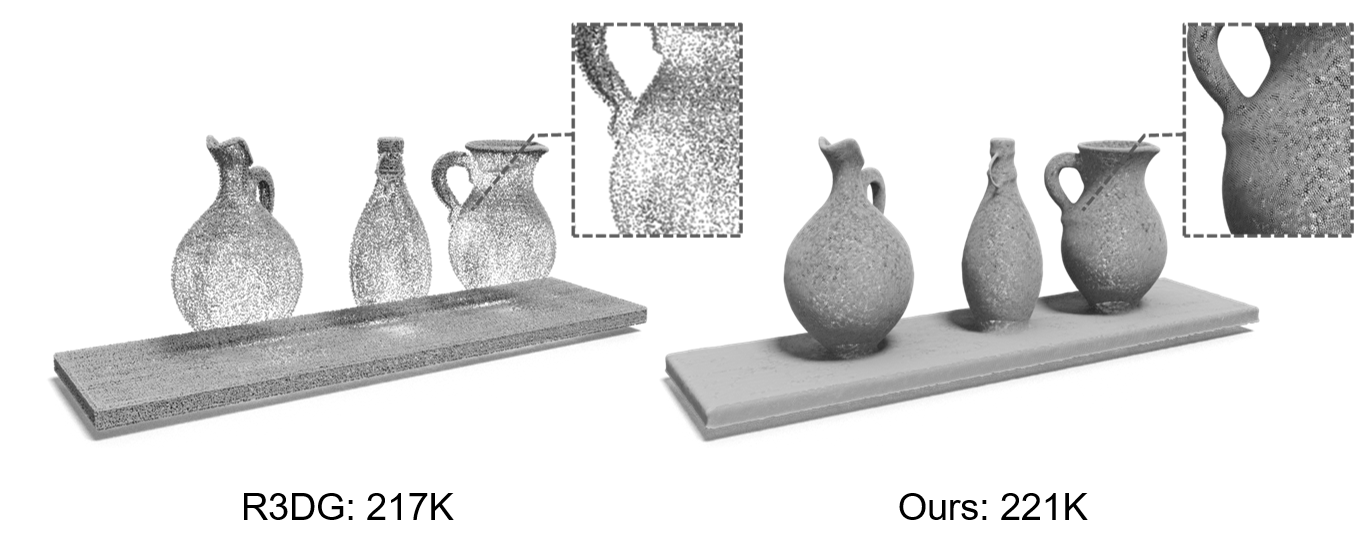}
  \end{center}
\vspace{-2em}
\caption{Comparison of points distribution between Ours and R3DG }
\label{fig:points}

\end{figure}

\begin{figure}[t!]
  \begin{center}

\includegraphics[width=0.5\textwidth]{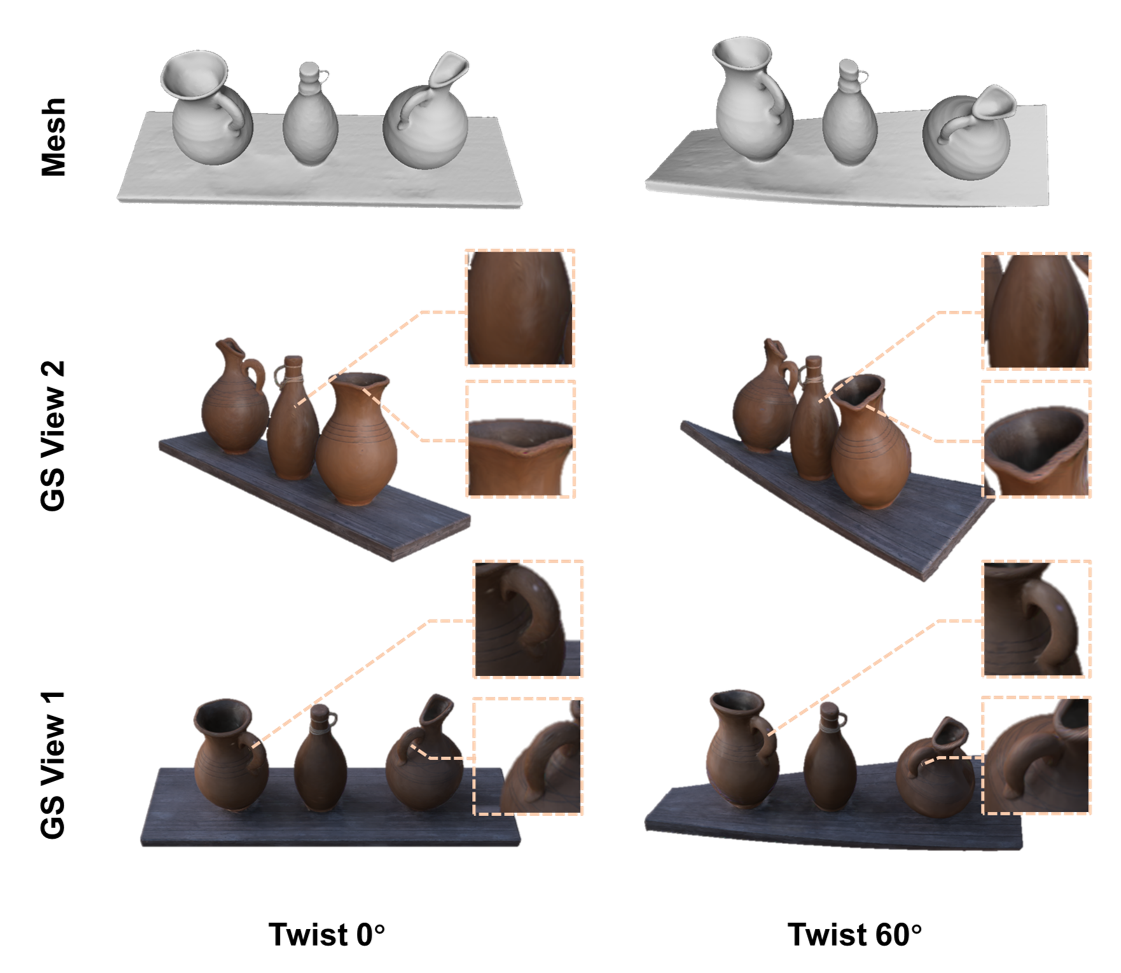}
  \end{center}
  \vspace{-2em}
\caption{GS relight with mesh deform }
\label{fig:deform}

\end{figure}

\noindent\textbf{Relighting Evaluation}
As presented in Tab. \ref{table:relight}, our version of initialization helps to improve relighting, albedo and roughness precision in the framework of R3DG. 
Benefiting from our Gaussian mesh binding method, we can effortlessly transfer parameters from R3DG to the mesh in a one-to-one correspondence. 
As demonstrated, the transferred metrics exhibit significant superiority in relighting over previous mesh-based approaches, while achieving these improvements with reduced computational time.
Qualitatively, our method achieves visually pleasing material decomposition, facilitating a realistic relighting effect (see Fig.~\ref{fig:relight}). 

Last but not the least, we visualize the distributions of Gaussian points in R3DG and those of our proposed method in Fig.~\ref{fig:points} -- With a comparable number of Gaussian points, the distribution in our method is explicitly guided by the underlying mesh geometry, resulting in a more uniform spatial allocation. 
This structural advantage directly contributes to the superior material learning performance of our approach compared to R3DG.

\noindent\textbf{Deformation Evaluation}
We validate the GS-mesh binding through large-scale geometric deformation, as visualized in Fig.\ref{fig:deform}. Applying a 60° X-axis twist to the jug mesh induces synchronized transformations on both the explicit surface and the bound Gaussians in R3DG. Crucially, the corresponding positional and normal adjustments of Gaussians preserve photorealistic interactions with environmental lighting: specular highlights shift coherently along the deformation path while cast shadows naturally elongate according to surface curvature changes. This parallel behavior of illumination effects demonstrates our method maintains physical consistency between mesh editing and GS manipulation. The results confirm that even under extreme topology changes, our binding mechanism successfully propagates deformations while retaining the original relighting properties of both representations.

\subsection{Ablation Study}
\begin{table}[!t]
    \centering
    \caption{Ablation Study on supervision and reprojection (DTU scenes)}
    \label{tab:ablation_dtu}
    \footnotesize
    \begin{tabular}{l|ccc|c}
    \toprule
    & \multicolumn{3}{c|}{\textbf{Rendering}} 
    & \multicolumn{1}{c}{\textbf{Geometry}} \\
    \cline{2-5}
    \textbf{Config} 
    & PSNR~$\uparrow$ & SSIM~$\uparrow$ & LPIPS~$\downarrow$ 
    & CD~$\downarrow$ \\
    \midrule
    Ours           
    & 26.21 & 0.906 & 0.148 
    & 0.70  \\
    w/o RGB Loss
    & 23.66 & 0.843 & 0.206
    & 0.89  \\
    w/o GEO Loss
    & 25.80 & 0.897 & 0.160
    & 0.73  \\
    w/o Length Control
    & 25.07 & 0.871 & 0.168
    & 0.71  \\
    \bottomrule
    \end{tabular}
\end{table}
In this section, we systematically evaluate the impact of specific components of our approach.

We start by analyzing the loss functions, which is demonstrated in Tab.~\ref{tab:ablation_dtu}. 
First, we discover that RGB loss supervision plays a critical role in our method, whose absence leads to a significant decrease in rendering quality and Chamfer Distance, highlighting its importance in capturing fine details and color information for accurate texture and geometric reconstruction. 
Second, the removal of geometry loss supervision results in a slight decrease in rendering quality and Chamfer Distance. 
Finally, omitting the edge length control based on texture density component also leads to a decrease in rendering quality, while Chamfer Distance remains stable. 
As illustrated in Fig.~\ref{fig:fft}, after incorporating our length control, the mesh demonstrates more detailed representations in texture-dense regions while retaining its original configuration in texture-uniform areas.

\begin{table}[!t]
    \centering
    \caption{Ablation Results on edge length initialization (scan 65 from DTU scenes)}
    \label{tab:pointsabl}
    \footnotesize
    \scalebox{0.92}{
    \begin{tabular}{l|ccc}
    \toprule
    & \multicolumn{3}{c}{(\textbf{Min Length, Max Length})} \\
    \cline{2-4}
    \textbf{Metrics} 
    & (\textbf{\(2 \mathrm{e}{-4}\), \(2 \mathrm{e}{-3}\)}) & (\textbf{\(4 \mathrm{e}{-4}\), \(4 \mathrm{e}{-3}\)}) & (\textbf{\(8 \mathrm{e}{-4}\), \(8 \mathrm{e}{-3}\)}) \\
    \midrule
    Chamfer Distance         
    & 0.76
    & 0.76
    & 0.82 \\
    Number of Vertices
    & 609K
    & 159K
    & 46K  \\
    \bottomrule
    \end{tabular}
    }
\end{table}

We further validate the edge-length control parameters by varying the minimum and maximum thresholds (Tab.~\ref{tab:pointsabl}). Decreasing the thresholds produces a much denser mesh (about \(4\times\)more vertices) but yields essentially no accuracy gain. In contrast, increasing the thresholds markedly reduces mesh resolution and leads to a clear drop in reconstruction quality. Overall, the mid-range setting \((4\mathrm{e}{-4},\,4\mathrm{e}{-3})\) provides a favorable trade-off between mesh complexity and reconstruction accuracy.



\section{Limitations and Conclusion}
We quantitatively observed that our refinement is less effective in scenarios with poor lighting conditions (see case 110 of Tab.\ref{table:dtugeo} and DutchOven in Tab.\ref{table:dtcgeo}).
We refer readers to Sec.~\ref{sup:failure} of the Supp. for details. 
This work presents a unified framework for jointly optimizing geometry and appearance, mitigating the geometry–texture misalignment in generic 3DGS pipelines. By co-optimizing mesh vertices and colors under photometric and geometric constraints, we produce high-fidelity, editable textured meshes. Coupling parametric Gaussians with mesh vertices further enables synchronized material control and surface deformation, improving reconstruction quality and supporting interactive 3D editing.
This advancement paves the way for more intuitive and efficient workflows in virtual environment design, digital content creation, and beyond, where cohesive geometry-appearance manipulation is essential. 

\paragraph*{Acknowledgements: }
This work is supported by the National Natural Science Foundation of China under contract No. 62171256, 62205178.

\small
\bibliographystyle{ieeenat_fullname}
\bibliography{references}

\clearpage
\setcounter{page}{1}
\maketitlesupplementary

\label{sec:appendix}

\section{Remesh Algorithm}\label{sup:remesh}
In this section, we provide a detailed description of the edge operations used in our mesh processing framework. Specifically, we discuss three fundamental operations: Edge Split, Edge Collapse, and Edge Flip, which allow for dynamic modification of the mesh topology while preserving geometric and color continuity. We demonstrated the detailed algorithm in Algorithm~\ref{alg:remesh}.

\begin{algorithm}[H]
\caption{Iterative Remeshing}
\label{alg:remesh}
\begin{algorithmic}[1]
\Require Mesh vertices and gradient features $\mathcal{V}_{etc} \in \mathbb{R}^{V \times D}$, vertex colors $\mathcal{C} \in \mathbb{R}^{V \times 3}$, vertex texture density $F_{l} \in \mathbb{R}^{V}$, faces $\mathcal{F} \in \mathbb{Z}^{F \times 3}$, edge length tolerance $\epsilon$, flip flag $\beta_{flip}$, color gradients $\nabla\mathcal{C}$, and max vertices $V_{max}$.
\Ensure Remeshed vertices $\mathcal{V}_{etc}$, colors $\mathcal{C}$, vertex texture density $F_{l}$, faces $\mathcal{F}$, and color gradients $\nabla\mathcal{C}$.
\begin{small}
    \State $L_{ref} \gets \mathcal{V}_{etc}[:, -1]$
    \State $L_{min} \gets L_{ref} \cdot (1-F_{l}) \cdot (1 - \epsilon)$ 
    \State $L_{max} \gets L_{ref} \cdot (1-F_{l}) \cdot (1 + \epsilon)$
    \State \textbf{--- Edge Collapse ---}
    \State $\mathcal{V} \gets \mathcal{V}_{etc}[:, :3]$
    \State $\mathcal{E}, \mathcal{F}_{E} \gets \text{CalculateEdges}(\mathcal{F})$
    \State $L_{E} \gets \text{CalculateEdgeLengths}(\mathcal{V}, \mathcal{E})$
    \State $\mathcal{P}_{collapse} \gets \text{CalculateFaceCollapses}(\mathcal{V}, \mathcal{F}, \mathcal{E}, \mathcal{F}_{E}, L_{E}, L_{min})$
    \State $S \gets \max(0, 1 - L_{E} / \text{mean}(L_{min}[\mathcal{E}]))$ \Comment{Shortness term}
    \State $\mathcal{P}_{priority} \gets \mathcal{P}_{collapse} + S$
    \State $\text{CollapseEdges}(\mathcal{V}_{etc}, \mathcal{C}, F_{l}, \mathcal{F}, \mathcal{E}, \mathcal{P}_{priority}, \nabla\mathcal{C})$

    \State \textbf{--- Edge Split ---}
    \If{$|\mathcal{V}_{etc}| < V_{max}$}
        \State $\mathcal{E}, \mathcal{F}_{E} \gets \text{CalculateEdges}(\mathcal{F})$
        \State $\mathcal{V} \gets \mathcal{V}_{etc}[:, :3]$
        \State $L_{E} \gets \text{CalculateEdgeLengths}(\mathcal{V}, \mathcal{E})$
        \State $\mathcal{S}_{split} \gets L_{E} > \text{mean}(L_{max}[\mathcal{E}])$
        \State $\text{SplitEdges}(\mathcal{V}_{etc}, \mathcal{C}, F_{l}, \mathcal{F}, \mathcal{E}, \mathcal{F}_{E}, \mathcal{S}_{split}, \nabla\mathcal{C})$
    \EndIf

    \State \textbf{--- Edge Flip ---}
    \State $\mathcal{V} \gets \mathcal{V}_{etc}[:, :3]$
    \If{$\beta_{flip}$}
        \State $\mathcal{E}, \_, \mathcal{E}_{\mathcal{F}} \gets \text{CalculateEdges}(\mathcal{F})$
        \State $\text{FlipEdges}(\mathcal{V}, \mathcal{F}, \mathcal{E}, \mathcal{E}_{\mathcal{F}})$
    \EndIf

\end{small}
\end{algorithmic}
\end{algorithm}

\section{Loss Function}\label{sup:loss}
In this section, we provide a detailed explanation of the loss functions used in our framework. These losses enforce photometric consistency, geometric accuracy, and regularization for stable optimization.

The RGB loss \( L_{rgb} \) is defined as:
\begin{equation}
\footnotesize
\mathcal{L}_{rgb} = \frac{1}{s} \sum_{i=1}^{s} \left( \alpha \| I_i - \hat{I}_i \| + (1 - \alpha)  \text{SSIM}(I_i, \hat{I}_i) \right)
\end{equation}
where \( I_i \) and \( \hat{I}_i \) are the predicted and ground-truth RGB images. \(\alpha\) is set to 0.8 following \cite{Kerbl20233DGS}.

The normal and depth map loss \(\mathcal{L}_{geo} \) can be defined as:
\begin{equation}
\footnotesize
\mathcal{L}_{geo} = \frac{1}{s} \sum_{i=1}^{s} \left( \| n_i - \hat{n}_i \| + \| d_i - \hat{d}_i \| \right)
\end{equation}
where \( n_i \) and \( \hat{n}_i \) are the predicted and pseudo-ground-truth normal maps, and \( d_i \) and \( \hat{d}_i \) are the predicted and pseudo-ground-truth normal maps, respectively.

The regularization loss \( \mathcal{L}_{reg} \) includes both Laplacian smoothing and mesh normal consistency:
\begin{equation}
\footnotesize
\mathcal{L}_{reg} = \frac{1}{n} \sum_{i=1}^{n} \| Lv_i \|^2 + \frac{1}{m} \sum_{i=1}^{m} \| N_i - \bar{N}_i \|^2
\end{equation}
where \( L \) is the Laplacian matrix,\(n\) is the number of vertices, \(m\) is the number of faces, \( v_i \) are the vertex positions, \( N_i \) are the face normals, and \( \bar{N}_i \) are the averaged normals of adjacent faces for mesh $\mathcal{M}$.

\section{Vertex-Gaussian Binding}\label{sup:gsbind}
1. \textbf{Position}: The position of each Gaussian is directly mapped from the vertex position in the mesh. Let \( \mu_i \in \mathbb{R}^3 \) represent the position of a guassian, then:
\[
\mu_i = v_i
\]

2. \textbf{Scale}: 
The scale of each Gaussian is represented as a vector \( \mathbf{s}_i = (s_1, s_2, s_3) \), where each component corresponds to different geometric properties of the mesh around the vertex \( v_i \). Specifically: \( s_2 \) is the length of the projection of the longest edge \( e_{\text{max}} \) onto the tangent plane at vertex \( v_i \). 

\( s_3 \) is the average projection length of all edges incident to vertex \( v_i \) onto the tangent plane.

Finally, \( s_1 \) is defined as the average of \( s_2 \) and \( s_3 \).
Thus, the scale vector \( \mathbf{s}_i \) for each Gaussian is composed of these three components \( s_1, s_2, s_3 \), reflecting both the local geometric properties of the vertex and the surrounding mesh structure.

3. \textbf{Rotation}:
The rotation matrix \( R_i \) for each Gaussian is determined by three orthogonal direction vectors: \( \mathbf{v}_1 \) is the normal vector at vertex \( v_i \), which is typically computed from the surrounding vertex neighbors and represents the direction perpendicular to the tangent plane at the vertex. \( \mathbf{v}_2 \) is the projection of the longest edge \( e_{\text{max}} \) onto the tangent plane at vertex \( v_i \). 

\( \mathbf{v}_3 \) is the vector that is orthogonal to both \( \mathbf{v}_1 \) and \( \mathbf{v}_2 \), ensuring the three vectors form an orthonormal basis. It can be computed as:
\(
\mathbf{v}_3 = \mathbf{v}_1 \times \mathbf{v}_2
\).

4. \textbf{Opacity}: In our method, we assign a constant opacity value of 0.9 to each Gaussian, assuming that every point on the mesh is visible.

5. \textbf{Spherical Harmonics (SH) coefficients}: In our method, we assign the low-order SH coefficients directly from the vertex color \( c_i \), and set the higher-order coefficients to zero.

\section{Failure Cases Analysis}\label{sup:failure}
\label{sup:failure}

\begin{figure}[t]
    \centering
    \includegraphics[width=\linewidth]{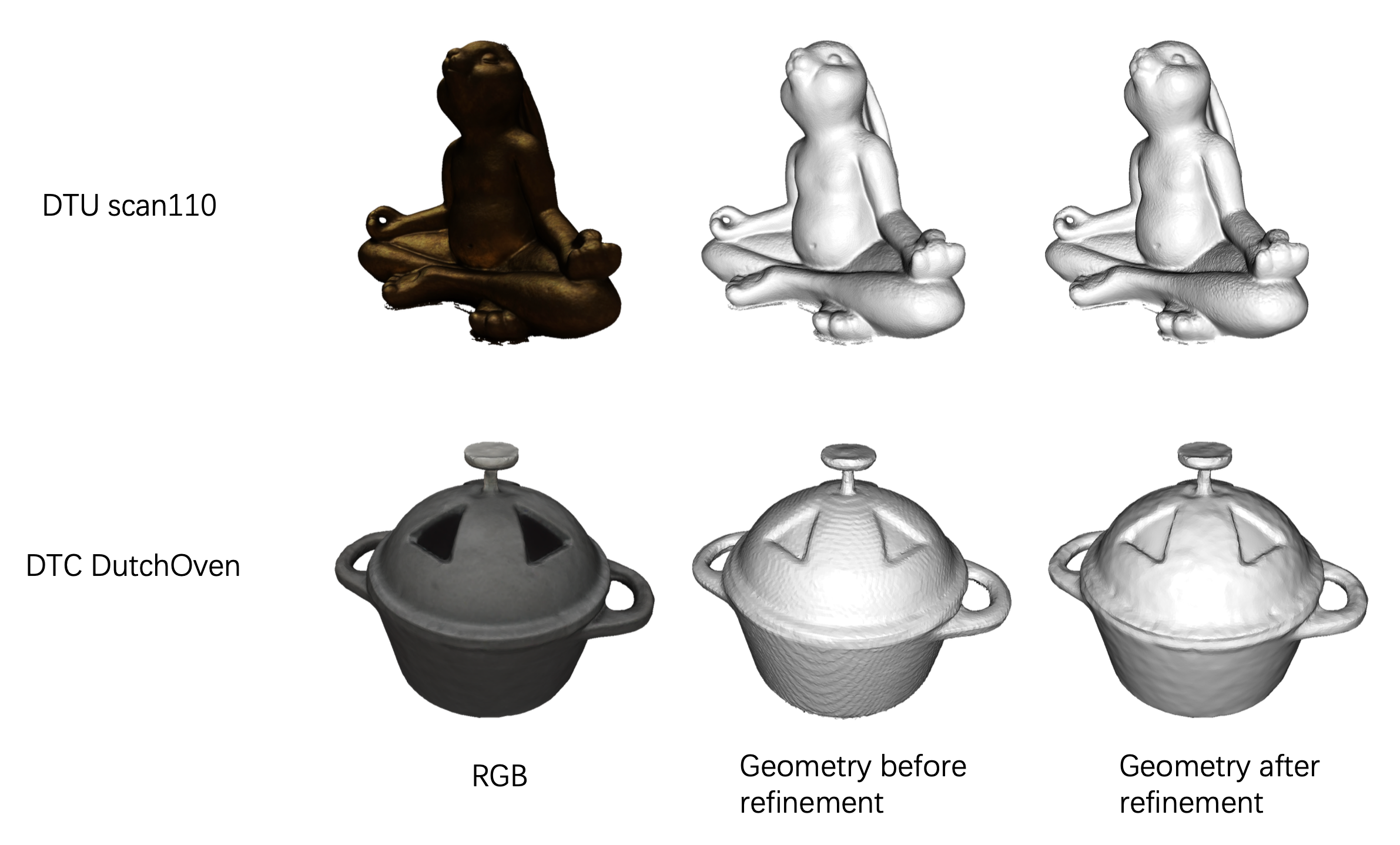}
    \caption{Visualization of failure cases under poor lighting conditions. Left: Case 110 from DTU dataset showing reconstruction artifacts in shadowed regions. Right: DutchOven from DTC dataset demonstrating degraded geometry in low-light condition.}
    \label{fig:failure}
\end{figure}

While our proposed refinement process consistently demonstrates enhancements in geometric accuracy and detail, its efficacy, like many state-of-the-art methods, is correlated with the quality of the photometric information in the input images. As illustrated in Fig.~\ref{fig:failure}, certain challenging lighting conditions, such as the presence of strong cast shadows or globally low-light environments, can present limitations. In these scenarios, the refinement may yield more subtle improvements or highlight areas for future research in robust reconstruction.

\subsection{Strong Shadows}

The "scan110" from the DTU dataset provides a valuable case study on the influence of high-contrast lighting. The input RGB image features a dark, specular object under strong directional light, resulting in areas of deep shadow. The initial geometry, shown as "Geometry before refinement," offers a coarse yet largely complete representation of the bunny figure.

Our refinement process demonstrates a clear benefit in the well-illuminated regions. On the figure's head and belly, for example, the surface is successfully smoothed, and details are sharpened, showcasing the method's effectiveness. In contrast, the regions obscured by shadow—specifically the lap, the underside of the chin, and between the limbs—present a more challenging scenario. The scarcity of reliable photometric cues in these areas makes it difficult for the algorithm, which leverages multi-view consistency, to resolve the geometry with the same level of confidence. This can lead to the introduction of localized surface artifacts. This observation suggests that integrating priors or specialized shadow-handling techniques could be a promising direction for future work to further enhance robustness in extreme lighting.

\subsection{Global Low-Light}

The "DutchOven" from the DTC dataset illustrates a different set of challenges associated with globally low-light conditions. Here, the input shows a dark, matte object with diffuse, dim illumination, leading to low contrast across the entire surface. The "Geometry before refinement" is of a modest quality, exhibiting a noisy surface where details, like the triangular patterns on the lid, are not yet fully resolved.

In this low signal-to-noise context, the refinement process achieves limited additional improvement over the initial geometry. As shown in "Geometry after refinement," the surface texture remains noisy, and the geometric details on the lid become less defined. This is because the refinement process finds it challenging to distinguish faint surface features from sensor noise in the low-contrast input images. This case highlights that a sufficient level of image quality and contrast is beneficial for achieving optimal results, a characteristic common to many photometric refinement techniques. It also suggests that our method could be further enhanced by coupling it with advanced image pre-processing, such as denoising or contrast enhancement, for inputs captured in such demanding conditions.

\end{document}